\documentclass[letterpaper, 10pt, conference]{ieeeconf}  

\IEEEoverridecommandlockouts                             
                                                          
\usepackage[T1]{fontenc}
\usepackage{aecompl}

\usepackage{ifpdf}
\usepackage{graphicx}
\usepackage{array}
\usepackage[caption=false,font=footnotesize]{subfig}
\usepackage{tabularx}
\usepackage{stfloats}
\usepackage{multirow}
\usepackage{algorithmic}
\usepackage{algorithm}

\usepackage{cite}
\usepackage{textcomp}

\usepackage{colortbl}  
\usepackage{verbatim}
\usepackage{url}
\usepackage{xcolor}
\usepackage{hyperref}
\usepackage{bbding}
\usepackage{booktabs}
\usepackage{easybmat}
\usepackage{blkarray}
\usepackage{enumerate}
\usepackage{pifont}
\usepackage{balance}

\usepackage{amsfonts}
\usepackage{amssymb}
\usepackage{amsmath}
\usepackage{mathtools}
\usepackage{arydshln} 

\newtheorem{lemma}{Lemma}

\renewcommand \arraystretch{1.2}

\DeclareMathOperator*{\Span}{Span}

\DeclareMathOperator{\col}{col}

\makeatletter
\renewcommand{\maketag@@@}[1]{\hbox{\m@th\normalsize\normalfont#1}}%
\makeatother

\usepackage{etoolbox}
\makeatletter
\patchcmd{\@makecaption}
  {\scshape}
  {}
  {}
  {}
\makeatletter
\patchcmd{\@makecaption}
  {\\}
  {.\ }
  {}
  {}
\makeatother

\title{\LARGE \bf
Preparation of Papers for IEEE Sponsored Conferences \& Symposia*
}

\title{KD-EKF: A Consistent Cooperative Localization Estimator Based on Kalman Decomposition}

\author{Ning Hao, Fenghua He, Chungeng Tian, Yu Yao, Weilong Xia
\thanks{N. Hao, F. He, C. Tian, Y. Yao and W. Xia are with the School of Astronautics, Harbin Insititute of Technology, Harbin 150001, China (email: haoning0082022@163.com; hefenghua@hit.edu.cn; 2018tian@gmail.com; yaoyu@hit.edu.cn; xiawl0524@gmail.com)}
}

\begin{document}

\maketitle
\thispagestyle{empty}
\pagestyle{empty}

\begin{abstract}
In this paper, we revisit the inconsistency problem of EKF-based cooperative localization (CL) from the perspective of system decomposition. By transforming the linearized system used by the standard EKF into its Kalman observable canonical form, the observable and unobservable components of the system are separated. Consequently, the factors causing the dimension reduction of the unobservable subspace are explicitly isolated in the state propagation and measurement Jacobians of the Kalman observable canonical form. Motivated by these insights, we propose a new CL algorithm called KD-EKF which aims to enhance consistency. The key idea behind the KD-EKF algorithm involves perform state estimation in the transformed coordinates so as to eliminate the influencing factors of observability in the Kalman observable canonical form. As a result, the KD-EKF algorithm ensures correct observability properties and consistency. We extensively  verify the effectiveness of the KD-EKF algorithm through both Monte Carlo simulations and real-world experiments. The results demonstrate that the KD-EKF  outperforms state-of-the-art algorithms in terms of accuracy and consistency.
\end{abstract}


\section{INTRODUCTION}

For a multi-robot system (MRS), determining each robot's pose consisting of position and attitude in a common reference frame is crucial for many missions, such as surveillance and reconnaissance \cite{A12}, search and rescue \cite{A1}, and formation control \cite{A4}. The most straightforward solution to localize an MRS is through a global positioning system (GPS or equivalent). Although this solution is easy to use, it might be unreliable or unavailable in some scenarios, such as underwater, underground, or indoors \cite{A37}. Alternatively, localizing a team of robots can be realized via feature matching with a prior map of the operating environment. This approach relies on environmental features and might not adapt to textureless places where finding landmarks is quite challenging. Consequently, a significant amount of research effort has been devoted to \emph{cooperative localization} (CL), i.e., robots cooperatively estimate their poses in a common reference frame by utilizing ego-motion information and relative robot-to-robot measurements \cite{A9}.

To solve CL problems, plenty of algorithms have been developed based on nonlinear estimation methodologies, including the extended Kalman filter (EKF) \cite{B25}, maximum a posteriori estimation \cite{B26}, and particle filter \cite{A28}, etc. Among these approaches, EKF is the most widely used because of its efficiency and accuracy. However, EKF suffers from inconsistency issues in CL \cite{B8}. A state estimator is called \emph{consistent} if the estimation errors are zero-mean and have covariance smaller than or equal to that calculated by the estimator \cite{B2}. Consistency is one of the primary criteria to evaluate the performance of estimators. If an estimator is inconsistent, the covariance calculated by the estimator tends to be overconfident, and thus cannot characterize the estimation uncertainty. Meanwhile, the overconfident covariance yields wrong information gain, which in turn affects the accuracy of state estimates \cite{B37}.

Observability is of great importance for consistent state estimation, which determines the minimal measurements required for the reconstruction of system states \cite{B36, B37}. Unluckily, the actual nonlinear CL system is not completely observable. Huang et. al., for the first time, discovered the connection between observability and consistency for both SLAM and CL problems \cite{B12, B14, B16, B8, B7, B37}. They proved  analytically that the linearized system employed by the standard EKF has an unobservable subspace of lower dimension than that of the underlying nonlinear system. The dimension reduction causes the estimator to erroneously gain information along the unobservable directions, thus leading to the estimator being inconsistent. In light of these in-depth analyses, first-estimates Jacobian (FEJ) \cite{B12, B8, B14} and observability-constrained (OC) \cite{B6, B13} methodologies were proposed to address this issue. The key idea behind these approaches is to guarantee correct observability properties by enforcing the unobservable subspace to span along proper directions. These algorithms have been verified to improve the consistency significantly in both SLAM and CL \cite{B13, B34}.

Inspired by these insightful works, we revisit the consistency problem of the standard EKF-CL from a new perspective, i.e., system decomposition. We introduce the Kalman observable decomposition which offers a procedure to present the CL system in a more illuminating form i.e., \emph{Kalman observable canonical form}. Specifically, we construct a time-varying transformation of coordinates by which the state space of the linearized system employed by the standard EKF is separated into observable and unobservable subspaces. More importantly, the two subspaces are orthogonal. Benefiting from this decomposition, the items leading to the dimension reduction in the unobservable subspace are explicitly isolated in the state propagation and measurement Jacobian matrices of the Kalman observable canonical form. This inspires us to perform state estimation in the transformed coordinates so that the inconsistency caused by the dimension mismatch of the unobservable subspace can be completely eliminated. In summary, the major contributions of this paper are as follows:
\begin{itemize}

\item We analyze the reasons leading to the dimension reduction in the unobservable subspace of the linearized system of the standard EKF and explicitly separate the influencing items causing such reduction.


%
\item We propose a novel CL algorithm, called Kalman decomposition-based EKF (KD-EKF), to improve the consistency and accuracy of CL. The proposed algorithm ensures both accurate and consistent estimation by precisely eliminating the influencing items of observability.


\item The KD-EKF CL algorithm is extensively validated in both Monte Carlo simulations and real-world experiments. It is shown that the proposed algorithm performs better than existing methods in terms of both accuracy and consistency.
\end{itemize}

\section{PROBLEM Formulation}
\label{sec:problem}
In this section, we briefly introduce the formulation of EKF-based cooperative localization (CL) which fuses ego-motion information of each robot and relative robot-to-robot measurements.

\subsection{System Model of CL}

Consider a team of $n$ robots jointly estimate their poses with respect to a common reference frame. Each robot is equipped with proprioceptive sensors to sense its ego-motion information, and exteroceptive sensors to measure relative robot-to-robot measurements.

\begin{figure}[!htp]
\centering
\includegraphics[scale=0.4]{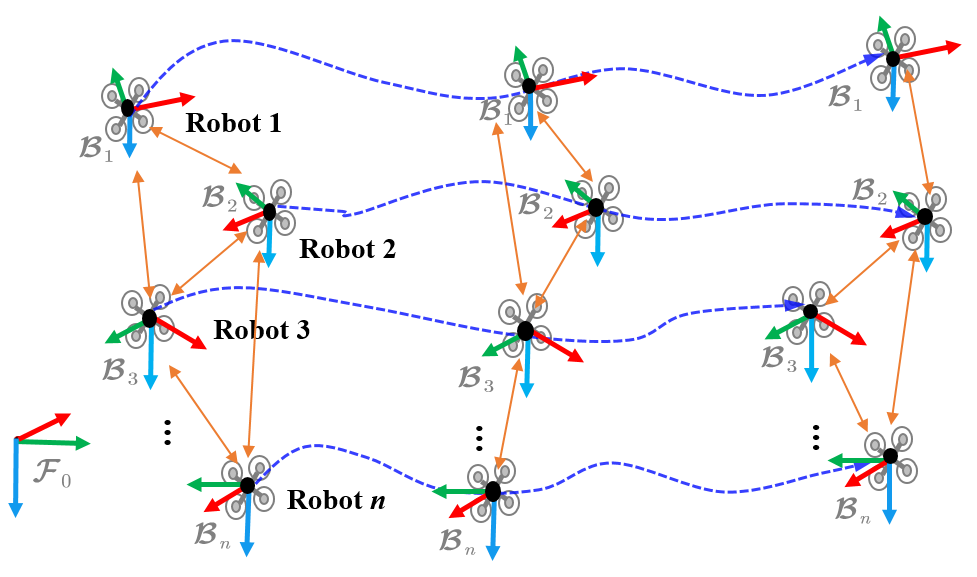}
\caption{Example scenario of a multi-robot system performing cooperative localization. Orange arrows indicate relative measurements among robots and blue dash lines represent the trajectories of each robot.}
\label{fig:ref}
\end{figure}

For filter design purposes, we formulate the localization problem in 2.5D as shown in Fig. \ref{fig:ref}, where the global reference frame $\mathcal{F}_0$ and the body-fixed reference frames $\mathcal{B}_i$ are chosen to be horizontal with their $z$-axis along the direction of gravity. In cooperative localization tasks, it is common practice to simplify the full 3D problem into 2.5D by taking the non-roll and non-pitch reference frames since the roll and pitch angles can be recovered from onboard IMUs \cite{A11}. Then only the position and orientation of the robots need to be determined.

Let $\boldsymbol{p}_{i, k} \in \mathbb{R}^3$ and $\psi_{i, k} \in \mathbb{R}$ denote the position and orientation of robot $i$ at instant $k$ expressed with respect to the global reference frame $\mathcal{F}_0$, respectively. Then the state vector of the $n$-robot system is
$$
\boldsymbol{x}_{k}
=
\left[
\begin{array}{ccc}
\boldsymbol{x}_{1, k}^{\top} & \cdots & \boldsymbol{x}_{n, k}^{\top}  \\
\end{array}
\right]^{\top}
$$
where 
$
\boldsymbol{x}_{i, k} 
= 
[
\setlength{\arraycolsep}{2.0pt}
\begin{array}{cc}
\boldsymbol{p}_{i, k}^{\top} & \psi_{i, k} \\
\end{array}
]^{\top}
$.
The discrete-time state equation of the $i$-th robot under consideration is
\begin{align} \label{equ:system_i_}
\boldsymbol{p}_{i,k} &= \boldsymbol{p}_{i,k-1} + \mathbf{C}(\psi_{i, k-1}) (\boldsymbol{v}_{i, k-1} + \boldsymbol{\nu}_{i, k-1}) \delta t \\
\psi_{i,k} &= \psi_{i,k-1} + (\omega_{i, k-1} + \varpi_{i, k-1}) \delta t 
\label{equ:system_i}
\end{align}
where $\boldsymbol{v}_{i, k} \in \mathbb{R}^3$ and $\omega_{i, k} \in \mathbb{R}$ denote the linear and angular velocities of robot $i$ expressed relative to the body-fixed reference frame $\mathcal{B}_i$, $\boldsymbol{n}_{i, k} = [\boldsymbol{\nu}_{i, k}^\top \ \varpi_{i, k}]^\top$ denotes zero-mean white Gaussian noises with covariance $\mathbf{Q}_{i, k}$, $\delta t$ is the sampling period and $\mathbf{C}(\psi_{i, k-1})$ is the elementary rotation matrix about $z$-axis of the angle $\psi_{i, k-1}$. The state model \eqref{equ:system_i_}-\eqref{equ:system_i} can be rewritten as the following nonlinear function
\begin{equation}
\begin{aligned}
\boldsymbol{x}_{i, k+1} 
= 
\mathbf{f}(\boldsymbol{x}_{i, k}, \boldsymbol{u}_{i, k}, \boldsymbol{n}_{i, k})
\end{aligned}
\label{equ:motion_equation}
\end{equation}
where $\boldsymbol{u}_{i, k} = [\boldsymbol{v}_{i, k}^{\top} \ \omega_{i, k}]^\top$. 


Let $\boldsymbol{y}_{ij, k} \in \mathbb{R}^3$ denote the relative position observation of robot $i$ with respect to robot ${j}$ at time $k$. The measurement equation is given by
\begin{equation}
\begin{aligned}
\boldsymbol{y}_{ij, k} 
&=
\mathbf{C}^{\top} (\psi_{i, k}) (\boldsymbol{p}_{j, k} - \boldsymbol{p}_{i, k}) + \boldsymbol{\eta}_{ij, k} \\
&\triangleq
\boldsymbol{h}_{ij}(\boldsymbol{x}_k) + \boldsymbol{\eta}_{ij,k} \\ 
\end{aligned}
\label{equ:relative_position}
\end{equation}
where the measurement noise $\boldsymbol{\eta}_{ij, k} \in \mathbb{R}^{3}$ is modeled as zero-mean Gaussian with $\boldsymbol{\eta}_{ij, k} \sim \mathcal{N}(\mathbf{0}, \mathbf{R}_{ij})$. The overall measurement vector of the $n$ robots are column stacked by that of each robot, i.e., 
$$
\boldsymbol{y}_k 
= 
\left[
\begin{array}{ccc}
\boldsymbol{y}_{1, k}^\top & \cdots & \boldsymbol{y}_{n, k}^\top 
\end{array}
\right]^\top
$$ 
where $\boldsymbol{y}_{i, k}$ denotes the relative measurements obtained by robot $i$ at time $k$ given by
$$
\boldsymbol{y}_{i, k} 
= 
\left[
\begin{array}{ccc}
\boldsymbol{y}_{i{i_1}, k}^\top & \cdots & \boldsymbol{y}_{i{i_{n-1}}, k}^\top 
\end{array}
\right]^\top
$$
with $i \in \{1, 2, \cdots, n\}$ and $i_\iota \in \{1, \cdots, n\} \backslash \{i\}$ with $\iota \in \{1, \cdots, {n-1} \}$. 
The entire measurement equation of the $n$-robot system can be written in a compact form as follows
\begin{equation}
\begin{aligned}
\boldsymbol{y}_k 
= 
\mathbf{h}(\boldsymbol{x}_k) + \boldsymbol{\eta}_k
\end{aligned}
\label{equ:measurement}
\end{equation}
where $\boldsymbol{\eta}_k$ is column stacked by $\boldsymbol{\eta}_{ij, k}$.


\subsection{Standard EKF-CL}
\label{subsec:problem_ekf}

Next, we present the propagation and update processes of the standard EKF-CL. During EKF propagation, each robot propagates the current state estimate $\hat{\boldsymbol{x}}_{i,k-1|k-1}$ forward using latest odometry readings to arrive at the prediction $\hat{\boldsymbol{x}}_{i, k|k-1}$. Linearizing (\ref{equ:motion_equation}) at $\hat{\boldsymbol{x}}_{i, {k-1|k-1}}$ yields the error-state propagation equation
$$
\tilde{\boldsymbol{x}}_{i, {k|k-1}}
=
\mathbf{F}_{i, {k-1}} \tilde{\boldsymbol{x}}_{i, {k-1|k-1}} + \mathbf{G}_{i, {k-1}} \boldsymbol{n}_{i, {k-1}} 
$$
where $\mathbf{F}_{i, k-1} \in \mathbb{R}^{4 \times 4}$ and $\mathbf{G}_{i, {k-1}} \in \mathbb{R}^{4 \times 4}$ are the state propagation Jacobian matrix and noise propagation matrix of robot $i$ given by
$$
\setlength{\arraycolsep}{2pt}
\begin{aligned}
\mathbf{F}_{i, k-1} 
&=
{\small
\left[
\begin{array}{cc}
\mathbf{I}_3             &   \mathbf{J} (\hat{\boldsymbol{p}}_{i, {k|k-1}} - \hat{\boldsymbol{p}}_{i, {k-1|k-1}})  \\ 
\mathbf{0}_{1 \times 3}  &   1  \\
\end{array}
\right] } \\
\end{aligned}
$$
$$
\setlength{\arraycolsep}{2pt}
\begin{aligned}
\mathbf{G}_{i, k-1} 
&=
{\small
\left[
\setlength{\arraycolsep}{3.0pt}
\begin{array}{cc}
\mathbf{C}(\hat{\psi}_{i, k-1|k-1}) \delta t & \mathbf{0}_{3 \times 1} \\
\mathbf{0}_{1 \times 3} & \delta t \\
\end{array}
\right] } \\
\end{aligned}
$$
with 
$
\mathbf{J}
=
{\scriptsize
\left[
\begin{array}{ccc}
0 & -1 & 0 \\
1 & 0  & 0 \\
0 & 0  & 0
\end{array}
\right] } .
$
By stacking all $n$ robots' error-states, we can obtain the error-state propagation equation of the entire CL system
\begin{equation}
\begin{aligned}
&\tilde{\mathbf{x}}_{k+1|k}  \\
&=  
{
\footnotesize
\left[
\setlength{\arraycolsep}{1.2pt}
\begin{array}{ccc}
\mathbf{F}_{1,k} & \cdots & \mathbf{0} \\   
\vdots &\ddots & \vdots  \\
\mathbf{0} & \cdots & \mathbf{F}_{n,k} \\
\end{array}
\right]  
\left[
\begin{array}{c}
\setlength{\arraycolsep}{1.2pt}
\tilde{\mathbf{x}}_{1, k|k}  \\
\vdots \\
\tilde{\mathbf{x}}_{n, k|k} \\
\end{array}
\right]
+
\left[
\setlength{\arraycolsep}{1.2pt}
\begin{array}{ccc}
\mathbf{G}_{1,k} & \cdots & \mathbf{0} \\   
\vdots &\ddots & \vdots  \\
\mathbf{0} & \cdots & \mathbf{G}_{n,k} \\
\end{array}
\right] 
\left[
\setlength{\arraycolsep}{1.2pt}
\begin{array}{c}
\mathbf{n}_{1, k} \\
\vdots \\
\mathbf{n}_{n, k} \\
\end{array}
\right] } \\
&\triangleq
\mathbf{F}_{k} \tilde{\mathbf{x}}_{k|k} + \mathbf{G}_{k} \mathbf{n}_{k}
\end{aligned}
\label{equ:ekf_state}
\end{equation}
where $\mathbf{F}_{k-1} \in \mathbb{R}^{4n \times 4n}$ and $\mathbf{G}_{k-1} \in \mathbb{R}^{4n \times 4n}$.

During EKF update step, the state prediction is fused with the latest measurements to arrive at a new estimate. By linearizing (\ref{equ:relative_position}) at current state prediction $\hat{\boldsymbol{x}}_{k|k-1}$, we have
$$
\tilde{\boldsymbol{y}}_{ij,k} = \mathbf{H}_{ij,k} \tilde{\boldsymbol{x}}_{k|k-1} + \boldsymbol{\eta}_{ij,k}
$$
where $\mathbf{H}_{ij,k} \in \mathbb{R}^{3 \times 4n}$ is the measurement Jacobian matrix
$$
\begin{aligned}
\mathbf{H}_{ij,k}
=
- \mathbf{C}^{\top}(\hat{\psi}_{i, {k|k-1}}) \mathbf{H}_{r_{ij,k}}
\end{aligned}
$$
with
$$
\mathbf{H}_{r_{ij,k}}
=
\begin{aligned}
\left[
\begin{array}{ccccccc}
\mathbf{0} & \cdots & \mathbf{H}_{r_{ij,k}}^{(i)} & \cdots & \mathbf{H}_{r_{ij,k}}^{(j)} &
\cdots & \mathbf{0}
\end{array}
\right]
\end{aligned}
$$
and
$$
\begin{aligned}
\mathbf{H}_{r_{ij,k}}^{(i)}
&= 
\left[
\begin{array}{cc}
\mathbf{I}_3 & \mathbf{J} (\hat{\boldsymbol{p}}_{j, {k|k-1}} - \hat{\boldsymbol{p}}_{i, {k|k-1}})   \\
\end{array} 
\right] \\
\mathbf{H}_{r_{ij,k}}^{(j)}
&= 
\left[
\begin{array}{cc}
-\mathbf{I}_3 & \mathbf{0}_{3 \times 1} \\
\end{array} 
\right]. \\
\end{aligned}
$$
By stacking $\mathbf{H}_{r_{ij,k}}$ for $j \in \{1, \cdots, n\} \backslash \{i\}$, we obtain the measurement Jacobian matrix of $\boldsymbol{y}_{i, k}$ as follows
$$
\begin{aligned}
\mathbf{H}_{i, k} 
&=
\boldsymbol{\Gamma}_{i, k}
{\small
\left[
\begin{array}{c}
\mathbf{H}_{r_{{ii_1},k}} \\
\vdots \\
\mathbf{H}_{r_{{ii_{n-1}},k}} \\
\end{array}
\right] } \triangleq
\boldsymbol{\Gamma}_{i, k} \mathbf{H}_{r_{i, k}} 
\end{aligned}
$$
where $\boldsymbol{\Gamma}_{i, k} = - \mathbf{I}_{n-1} \otimes \mathbf{C}^{\top}(\hat{\psi}_{i, {k|k-1}}) $ with $\otimes$ standing for Kronecker product. The measurement Jacobian matrix of the entire CL system is a block matrix comprising of $\mathbf{H}_{i, k}$ for $i \in \{1, \cdots, n\}$, i.e.,
$$
\begin{aligned}
\mathbf{H}_k 
&=
{\small
\left[
\begin{array}{ccc}
\boldsymbol{\Gamma}_{1, k} & \cdots & \mathbf{0} \\
\vdots & \ddots & \vdots \\
\mathbf{0} & \cdots & \boldsymbol{\Gamma}_{n, k} \\
\end{array}
\right]
\left[
\begin{array}{c}
\mathbf{H}_{r_{1, k}} \\
\vdots \\
\mathbf{H}_{r_{n, k}} \\
\end{array}
\right] } \\
&\triangleq
\boldsymbol{\Gamma}_k \mathbf{H}_{r_k} .
\end{aligned}
$$
The measurement-error equation of the entire CL system can be written as
\begin{equation}
\tilde{\boldsymbol{y}}_{k} = \mathbf{H}_{k} \tilde{\boldsymbol{x}}_{k|k-1} + \boldsymbol{\eta}_{k} .
\label{equ:ekf_update}
\end{equation}
Then, the linearized error-state system model composed of \eqref{equ:ekf_state} and \eqref{equ:ekf_update} is employed by the standard EKF for state estimation.

\subsection{Observability Properties of CL}

The observability of a system plays a significant role in state estimation, which measures how well the states of a system can be inferred from knowledge of measurements. Since the absolute position and orientation information are unavailable, the $n$-robot cooperative localization system under consideration is not observable. 

In particular, the observability matrix for the linearized system employed by the standard EKF over the time interval $[k, \ k+\ell]$ is defined by
\begin{equation}
\mathcal{O}_k
\triangleq
{\small
\left[
\begin{array}{c}
\mathbf{H}_{k} \\
\mathbf{H}_{k+1} \mathbf{F}_{k} \\
\vdots \\
\mathbf{H}_{k+\ell} \mathbf{F}_{k+\ell-1} \cdots \mathbf{F}_{k} \\
\end{array}
\right]}.
\label{equ:obs_mat}
\end{equation}
The observability matrix's null space describes the corresponding system's unobservable subspace. It was demonstrated in \cite{B17} that the null space of $\mathcal{O}_k$ should span along the following four directions
\begin{equation}
\mathbf{N}^{\boldsymbol{x}}_k
=
{
\small
\Span_{\col}
\left[
\begin{array}{cc}
\mathbf{J} \boldsymbol{p}_{1, {k}} & \mathbf{I}_{2} \\
1 & \mathbf{0}_{1 \times 2}  \\
\vdots & \vdots \\
\mathbf{J}\boldsymbol{p}_{n, {k}} & \mathbf{I}_{2} \\
1 & \mathbf{0}_{1 \times 2} \\
\end{array}
\right] }
\label{equ:N_x}
\end{equation}
where the first column indicates global orientation and the second block column represents global position.

Note that the null space $\mathbf{N}^{\boldsymbol{x}}_k$ given in \eqref{equ:N_x} describes the unobservable directions of the underlying nonlinear CL system. When designing a consistent estimator for CL, we would like the system model employed by the estimator to have an unobservable subspace spanned by these directions. However, this is not the case for the standard EKF. The linearized system employed by the standard EKF, which is linearized about the current best state estimates, has one fewer unobservable direction. In particular, the global orientation erroneously becomes observable. This will lead the estimator to surreptitiously gain spurious information along the direction of global orientation, and eventually inconsistent \cite{B7}.

\section{System Decomposition} 
\label{sec:consistency}

In this section, we revisit the inconsistency problem from the view of system decomposition. By using Kalman observable decomposition, we separate the observable and unobservable components of the linearized system employed by the standard EKF. In the new system representation, the items leading to the erroneous shrinking of the unobservable subspace are recognized and isolated. Therefore, when designing consistent CL algorithms, correct observability properties can be maintained by explicitly eliminating these influencing items of observability.

\subsection{Kalman Observable Decomposition}
\label{subsec:ekf_consistency}

Kalman observable decomposition offers a non-singular coordinate transformation to transform a linear system into a Kalman observable canonical form \cite{B10}. In the Kalman observable canonical form, the entire state vector is decomposed into two \emph{orthogonal} components, i.e., the observable state and the unobservable state. Benefiting from the decomposition, the observable and unobservable states are completely separated. This allows us to analyze the influencing factors of observability more easily.


Hereafter, we construct a coordinate transformation matrix to transform the linearized system \eqref{equ:ekf_state}-\eqref{equ:ekf_update} employed by the standard EKF into its Kalman observable canonical form. 

\begin{lemma}
By using the following non-singular coordinate transformation matrix 
\begin{equation}
\mathbf{T}_{k} =
\begin{footnotesize}
\left[
\setlength{\arraycolsep}{1.8pt}
\begin{array}{cc|ccccc}
\mathbf{I}_3 & \mathbf{J} (\hat{\boldsymbol{p}}_{2, {k|k}} - \hat{\boldsymbol{p}}_{1, {k|k}}) & -\mathbf{I}_3 & \mathbf{0}_{3 \times 1} & \cdots & \mathbf{0}_{3 \times 3} & \mathbf{0}_{3 \times 1} \\
\mathbf{0}_{1 \times 3} & -1 & \mathbf{0}_{1 \times 3} & 1 & \cdots & \mathbf{0}_{1 \times 3} & 0  \\
\vdots & \vdots & \vdots & \vdots & \ddots & \vdots & \vdots  \\
\mathbf{I}_3 & \mathbf{J} (\hat{\boldsymbol{p}}_{n, {k|k}} - \hat{\boldsymbol{p}}_{1, {k|k}}) & \mathbf{0}_{3 \times 3} & \mathbf{0}_{3 \times 1} & \cdots & -\mathbf{I}_3 & \mathbf{0}_{3 \times 1}  \\
\mathbf{0}_{1 \times 3} & -1 &  \mathbf{0}_{1 \times 3} & 0 & \cdots & \mathbf{0}_{1 \times 3} & 1 \\
\hline
\mathbf{0}_{1 \times 3} & 1 & \mathbf{0}_{1 \times 3} & 0 & \cdots & \mathbf{0}_{1 \times 3} & 0 \\
\mathbf{I}_3 & \mathbf{0}_{3 \times 1} & \mathbf{0}_{3 \times 3} & \mathbf{0}_{3 \times 1} & \cdots & \mathbf{0}_{3 \times 3} & \mathbf{0}_{3 \times 1} \\
\end{array}
\right],
\end{footnotesize}
\label{equ:T_ekf}
\end{equation}
the linearized system \eqref{equ:ekf_state}-\eqref{equ:ekf_update} employed by the standard EKF can be transformed into the Kalman observable canonical form
\begin{align} \label{equ:canonical_ekf_f}
\tilde{\boldsymbol{z}}_{{k|k-1}} 
&= 
\boldsymbol{\mathcal{F}}_{{k-1}} \tilde{\boldsymbol{z}}_{{k-1|k-1}} 
+ 
\boldsymbol{\mathcal{G}}_{k-1} \boldsymbol{n}_{k-1} \\
\tilde{\boldsymbol{y}}_k \quad
&=
\boldsymbol{\mathcal{H}}_{k} \tilde{\boldsymbol{z}}_{{k|k-1}}
+ 
\boldsymbol{\eta}_{k} \label{equ:canonical_ekf_h}
\end{align}
where $\tilde{\boldsymbol{z}} \in \mathbb{R}^{4n}$ denotes the converted error-state expressed in the transformed coordinates,
the transformed noise propagation matrix is given by $\boldsymbol{\mathcal{G}}_{k-1} = \mathbf{T}_{k}\mathbf{G}_{k-1}$, and the transformed state propagation Jacobian matrix $\boldsymbol{\mathcal{F}}_{k-1} \in \mathbb{R}^{4n \times 4n}$ and measurement Jacobian matrix $\boldsymbol{\mathcal{H}}_{k} \in \mathbb{R}^{3m \times 4n}$ are
\begin{equation}
\begin{aligned}
\small
\boldsymbol{\mathcal{F}}_{{k-1}}
&= 
\begin{scriptsize}
\left[
\setlength{\arraycolsep}{1.0pt}
\begin{array}{ccccc|cc}
\mathbf{I}_3 & -\mathbf{J} \boldsymbol{\delta}_{2, k}  & \quad \cdots & \quad \mathbf{0}_{3 \times 3} & \mathbf{0}_{3 \times 1} & \cellcolor{blue!10} \boldsymbol{\Delta}_{21, k} & \cellcolor{green!30} \mathbf{0}_{3 \times 3}  \\ 
\mathbf{0}_{1 \times 3} & 1 & \quad \cdots & \quad \mathbf{0}_{1 \times 3} & 0 & \cellcolor{blue!10} 0 & \cellcolor{green!30} \mathbf{0}_{1 \times 3} \\
\vdots & \vdots & \quad \ddots & \quad \vdots & \vdots & \cellcolor{blue!10}  \vdots & \cellcolor{green!30} \vdots  \\
\mathbf{0}_{3 \times 3} & \mathbf{0}_{3 \times 1} & \quad \cdots & \quad  \mathbf{I}_3 & -\mathbf{J} \boldsymbol{\delta}_{n, k} & \cellcolor{blue!10} \boldsymbol{\Delta}_{n1, k} & \cellcolor{green!30} \mathbf{0}_{3 \times 3} \\
\mathbf{0}_{1 \times 3} & 0 & \quad \cdots & \quad \mathbf{0}_{1 \times 3} & 1 & \cellcolor{blue!10} 0 & \cellcolor{green!30} \mathbf{0}_{1 \times 3} \\
\hline
\mathbf{0}_{1 \times 3} & 0 & \quad \cdots & \quad \mathbf{0}_{1 \times 3} & 0 & \cellcolor{red!15} 1 & \cellcolor{red!15} \mathbf{0}_{1 \times 3} \\
\mathbf{0}_{3 \times 3} & \mathbf{0}_{3 \times 1} & \quad \cdots & \quad \mathbf{0}_{3 \times 3} & \mathbf{0}_{3 \times 1} & \cellcolor{red!15} \mathbf{J} \boldsymbol{\delta}_{1, k} & \cellcolor{red!15} \mathbf{I}_3 \\
\end{array}
\right]
\end{scriptsize}
\end{aligned}
\label{equ:canonical_ekf_state}
\end{equation}
\begin{equation}
\begin{aligned}
\small
\boldsymbol{\mathcal{H}}_{k}
&=
\begin{scriptsize}
\boldsymbol{\Gamma}_k
\left[
\setlength{\arraycolsep}{0.5pt}
\begin{array}{ccccccc|cc}
\mathbf{I}_3 & \mathbf{0}_{3 \times 1} & \mathbf{0}_{3 \times 3} & \mathbf{0}_{3 \times 1} & \cdots & \mathbf{0}_{3 \times 3} & \mathbf{0}_{3 \times 1} & \cellcolor{blue!10}  \boldsymbol{\Delta}_{21, k} & \cellcolor{green!30} \mathbf{0}_{3 \times 3} \\
\mathbf{0}_{3 \times 3} & \mathbf{0}_{3 \times 1} & \mathbf{I}_3 & \mathbf{0}_{3 \times 1} & & & & \cellcolor{blue!10} \boldsymbol{\Delta}_{31, k} & \cellcolor{green!30} \mathbf{0}_{3 \times 3} \\ 
\vdots & \vdots & & & \ddots & & & \cellcolor{blue!10} \vdots & \cellcolor{green!30} \vdots \\ 
\mathbf{0}_{3 \times 3} & \mathbf{0}_{3 \times 1} & & & & \mathbf{I}_3 & \mathbf{0}_{3 \times 1} & \cellcolor{blue!10} \boldsymbol{\Delta}_{n1, k} & \cellcolor{green!30} \mathbf{0}_{3 \times 3} \\
\hdashline
-\mathbf{I}_3 & \mathbf{J} \boldsymbol{\delta}_{12, {k}} & \mathbf{0}_{3 \times 3} & \mathbf{0}_{3 \times 1} & \cdots & \mathbf{0}_{3 \times 3} & \mathbf{0}_{3 \times 1 } & \cellcolor{blue!10} \boldsymbol{\Delta}_{12, k} & \cellcolor{green!30} \mathbf{0}_{3 \times 3} \\ 
-\mathbf{I}_3 & \mathbf{J} \boldsymbol{\delta}_{32, {k}} & \mathbf{I}_3 & \mathbf{0}_{3 \times 1} & & & & \cellcolor{blue!10} \boldsymbol{\Delta}_{32, k} & \cellcolor{green!30} \mathbf{0}_{3 \times 3} \\
\vdots & \vdots & & & \ddots & & & \cellcolor{blue!10}  \vdots & \cellcolor{green!30} \vdots \\ 
-\mathbf{I}_3 & \mathbf{J} \boldsymbol{\delta}_{n2, {k}} & & & & \mathbf{I}_3 & \mathbf{0}_{3 \times 1} & \cellcolor{blue!10} \boldsymbol{\Delta}_{n2, k} & \cellcolor{green!30} \mathbf{0}_{3 \times 3} \\
\hdashline
\multicolumn{7}{c|}{\vdots} & \cellcolor{blue!10}  \vdots &  \cellcolor{green!30} \vdots \\
\hdashline 
\mathbf{0}_{3 \times 3} & \mathbf{0}_{3 \times 1} & \cdots & \mathbf{0}_{3 \times 3} & \mathbf{0}_{3 \times 1} & -\mathbf{I}_3 & \mathbf{J} \boldsymbol{\delta}_{1n, {k}} & \cellcolor{blue!10} \boldsymbol{\Delta}_{1n, k} & \cellcolor{green!30} \mathbf{0}_{3 \times 3} \\ 
\mathbf{I}_3 & \mathbf{0}_{3 \times 1} & & & & -\mathbf{I}_3 & \mathbf{J} \boldsymbol{\delta}_{2n, {k}} & \cellcolor{blue!10} \boldsymbol{\Delta}_{2n, k} & \cellcolor{green!30} \mathbf{0}_{3 \times 3} \\ 
& & \ddots & & & \vdots & \vdots & \cellcolor{blue!10}  \vdots & \cellcolor{green!30} \vdots \\ 
& & & \mathbf{I}_3 & \mathbf{0}_{3 \times 1} & -\mathbf{I}_3 & \mathbf{J} \boldsymbol{\delta}_{{n-1n}, {k}} & \cellcolor{blue!10} \boldsymbol{\Delta}_{n-1n, k} & \cellcolor{green!30} \mathbf{0}_{3 \times 3} \\ 
\end{array}
\right]
\end{scriptsize}
\end{aligned}
\label{equ:canonical_ekf_measurement}
\end{equation}
with
$$
\begin{aligned}
\boldsymbol{\delta}_{i, k} = \hat{\boldsymbol{p}}_{i, {k|k-1}} - \hat{\boldsymbol{p}}_{i, {k-1|k-1}}, \\
\boldsymbol{\delta}_{ij, {k}} = \hat{\boldsymbol{p}}_{i, {k|k-1}} - \hat{\boldsymbol{p}}_{j, {k|k-1}},
\end{aligned}
$$
\begin{equation}
\mathbf{\Delta}_{ij, k} = \mathbf{J} ((\hat{\boldsymbol{p}}_{i, {k|k-1}} - \hat{\boldsymbol{p}}_{i, {k|k}}) - (\hat{\boldsymbol{p}}_{j, {k|k-1}} - \hat{\boldsymbol{p}}_{j, {k|k}}))
\label{equ:ekf_a_ij}
\end{equation}
for $i, j \in \{1, \cdots, n\}$. 
\label{lemma:std_ekf_dec}
\end{lemma}

\begin{proof}
Using the coordinate transformation defined in \eqref{equ:T_ekf}, the error-state can be converted into the transformed coordinates
\begin{align}
\tilde{\boldsymbol{z}}_{k-1|k-1} &= \mathbf{T}_{k-1} \tilde{\boldsymbol{x}}_{k-1|k-1} 
\label{equ:state_transform_k-1} \\
\tilde{\boldsymbol{z}}_{k|k-1} &= \mathbf{T}_{k} \tilde{\boldsymbol{x}}_{k|k-1}
\label{equ:state_transform_k}
\end{align}
where $\tilde{\boldsymbol{z}}_{k-1|k-1}$ and $\tilde{\boldsymbol{z}}_{k|k-1}$ denote the estimated and predicted error-state expressed in the transformed coordinates. Substituting \eqref{equ:state_transform_k-1} and \eqref{equ:state_transform_k} into the error-state propagation equation \eqref{equ:ekf_state}, we have
\begin{equation}
\begin{aligned}
\tilde{\boldsymbol{z}}_{k|k-1} &= 
\mathbf{T}_{k} \mathbf{F}_{k-1} \mathbf{T}_{k-1}^{-1} \tilde{\boldsymbol{z}}_{k-1|k-1} + \mathbf{T}_{k} \mathbf{G}_{k-1} \boldsymbol{n}_{k-1} \\
&\triangleq
\boldsymbol{\mathcal{F}}_{k-1} \tilde{\boldsymbol{z}}_{k-1|k-1} + \boldsymbol{\mathcal{G}}_{k-1} \boldsymbol{n}_{k-1}  .
\end{aligned}
\label{equ:trans_state_jac}
\end{equation}
Likewise, substituting \eqref{equ:state_transform_k} into the measurement-error equation \eqref{equ:ekf_update} yields
\begin{equation}
\begin{aligned}
\tilde{\boldsymbol{y}}_{k} 
&=  
\mathbf{H}_{k} \mathbf{T}_{k}^{-1}  \tilde{\boldsymbol{z}}_{k|k-1} + \boldsymbol{\eta}_{k} \\ 
&\triangleq
\boldsymbol{\mathcal{H}}_{k}  \tilde{\boldsymbol{z}}_{k|k-1} + \boldsymbol{\eta}_{k} .
\end{aligned}
\label{equ:trans_meas_jac}
\end{equation}
In terms of \eqref{equ:trans_state_jac} and \eqref{equ:trans_meas_jac} as well as \eqref{equ:T_ekf}, \eqref{equ:canonical_ekf_state} and \eqref{equ:canonical_ekf_measurement} are easily obtained, which completes the proof.

\end{proof}

\subsection{Observability Analysis}

Using the non-singular coordinate transformation \eqref{equ:T_ekf}, the entire state vector is separated into observable and unobservable components. In particular, the four degree of freedom unobservable states, i.e., the global position and orientation, are explicitly expressed in the transformed coordinates. For clarity, the block matrices corresponding to the four unobservable states are marked with colors in the transformed state propagation Jacobian matrix \eqref{equ:canonical_ekf_state} and the transformed measurement Jacobian matrix \eqref{equ:canonical_ekf_measurement}. We should point out that in \eqref{equ:canonical_ekf_state} and \eqref{equ:canonical_ekf_measurement}, the block matrices colored by green corresponds to the global position, and the block matrices colored by purple corresponds to the global orientation. 

If the linearized system used by the standard EKF has the expected observability properties described by \eqref{equ:N_x}, both the purple and green block matrices in \eqref{equ:canonical_ekf_state} and \eqref{equ:canonical_ekf_measurement} should be zeros. Observe the block matrices in \eqref{equ:canonical_ekf_state} and \eqref{equ:canonical_ekf_measurement} colored by green, and all entries equal to zero. Therefore, the global position is not observable. However, in the purple area, the items $\boldsymbol{\Delta}_{ij, k}$ are not equal to zero since the state corrections of different robots are random and generally unequal (see \eqref{equ:ekf_a_ij}). This destroys the observability properties of the linearized system used by the standard EKF. In particular, the global orientation becomes observable. 

Due to the existence of the non-zero items $\boldsymbol{\Delta}_{ij, k}$ in \eqref{equ:canonical_ekf_state} and \eqref{equ:canonical_ekf_measurement}, the unobservable subspace of the linearized system used by the standard EKF is erroneously reduced by one along the global orientation. As a result, the standard EKF underestimates the uncertainty of the orientation estimates, and the estimator will be inconsistent. We note that the linearized system will have the expected observability properties if the entries $\boldsymbol{\Delta}_{ij, k}$ are annihilated. More importantly, the items $\boldsymbol{\Delta}_{ij, k}$, which are determined by the state corrections of the estimator, are rather small. Thus, the linearization errors introduced by eliminating these items can be negligible. This motivates the consistency improvement CL algorithm in the subsequent section.

\section{Kalman Decomposition Based Cooperative Localization Algorithm}
\label{sec:algorithm}

In the preceding section, we have identified and separated the items that lead to the dimension reduction of the unobservable subspace of the linearized system. Inspired by the analysis, we propose a CL estimator called KD-EKF to achieve consistent estimation. The key idea is to perform state estimation in the transformed coordinates to eliminate the items causing inconsistency. 



We now present the main procedures of the KD-EKF CL algorithm which are summarized as follows:
\subsubsection{Initialization}
During this stage, we first initialize the state $\boldsymbol{\hat{x}}_{0|0}$ and covariance $\mathbf{P}_{0|0}$ using the priori pose and uncertainty. Then we transform the initial covariance into the transformed coordinates by
\begin{equation}
\boldsymbol{\mathcal{P}}_{0|0} = \mathbf{T}_0 \mathbf{P}_{0|0} \mathbf{T}_0^{\top}
\end{equation}
where the coordinate transformation matrix $\mathbf{T}_0$ is obtained according to \eqref{equ:T_ekf}. The initialization step is only implemented at the start of state estimation. After initialization, the remaining four steps are implemented in an iterative manner.

\subsubsection{Propagation}
At this step, we propagate the state estimates of each robot via the nonlinear state model \eqref{equ:motion_equation}:
\begin{equation}
\hat{\boldsymbol{x}}_{i, k|k-1} = \mathbf{f}(\hat{\boldsymbol{x}}_{i, k-1|k-1}, \boldsymbol{u}_{i, k-1}, \mathbf{0})
\label{equ:state_propagation}
\end{equation}
where the noise vector $\mathbf{n}_{i, k-1}$ is set to zero.

\subsubsection{Jacobians computation}
At this step, we will derive the observable canonical form used by the KD-EKF for state estimation. By annihilating the items $\boldsymbol{\Delta}_{ij, k}$ in \eqref{equ:canonical_ekf_state} and \eqref{equ:canonical_ekf_measurement}, the to-be-used Kalman observable canonical form can be written as follows
\begin{align} \label{equ:auxiliary_system_f}
\tilde{\boldsymbol{z}}_{k|k-1} &= \bar{\boldsymbol{\mathcal{F}}}_{k-1} \tilde{\boldsymbol{z}}_{k-1|k-1} +  \bar{\boldsymbol{\mathcal{G}}}_{k-1} \boldsymbol{n}_{k-1} \\
\tilde{\boldsymbol{y}}_{k} \quad &= \bar{\boldsymbol{\mathcal{H}}}_{k} \tilde{\boldsymbol{z}}_{k|k-1} + \boldsymbol{\eta}_{k} \label{equ:auxiliary_system_h}
\end{align}
where the transformed state propagation and measurement Jacobian matrix are given by
\begin{align}
\bar{\boldsymbol{\mathcal{F}}}_{k-1}
&=
\left. \boldsymbol{\mathcal{F}}_{k-1} \right|_{\boldsymbol{\Delta}_{ij, k} = \mathbf{0}} 
\label{equ:aux_state_jac} \\
\bar{\boldsymbol{\mathcal{H}}}_{k} 
&= 
\left. \boldsymbol{\mathcal{H}}_{k} \right|_{\boldsymbol{\Delta}_{ij, k} = \mathbf{0}} 
\label{equ:aux_measure_jac} 
\end{align}
and the transformed noise propagation Jacobian matrix is computed via 
\begin{equation}
\left. \bar{\boldsymbol{\mathcal{G}}}_{k-1} = \boldsymbol{\mathcal{G}}_{k-1} \right|_{\hat{\boldsymbol{x}}_{k|k} = \hat{\boldsymbol{x}}_{k|k-1}}.
\label{equ:aux_noise_jac}
\end{equation}

\subsubsection{Update}
At this step, we are ready to apply the Kalman observable canonical form \eqref{equ:auxiliary_system_f}-\eqref{equ:auxiliary_system_h} for error-state and covariance estimation. Following the procedure of the standard EKF, the covariance is propagated via
\begin{equation}
\boldsymbol{\mathcal{P}}_{k|k-1} = \bar{\boldsymbol{\mathcal{F}}}_{k-1} \boldsymbol{\mathcal{P}}_{k-1|k-1} \bar{\boldsymbol{\mathcal{F}}}_{k-1}^{\top} + \bar{\boldsymbol{\mathcal{G}}}_{k-1} \mathbf{Q}_{k-1} \bar{\boldsymbol{\mathcal{G}}}_{k-1}^{\top} .
\label{equ:alg_cov_propagate}
\end{equation}
Then the covariance matrix is updated according to
\begin{equation}
\begin{aligned}
\boldsymbol{\mathcal{P}}_{k|k} 
= 
(\mathbf{I} - \boldsymbol{\mathcal{K}}_{k} \bar{\boldsymbol{\mathcal{H}}}_{k}) \boldsymbol{\mathcal{P}}_{k|k-1} 
\end{aligned}
\label{equ:alg_cov_update}
\end{equation}
where the information gain matrix $\boldsymbol{\mathcal{K}}_{k}$ is calculated by
\begin{equation}
\boldsymbol{\mathcal{K}}_{k} = \boldsymbol{\mathcal{P}}_{k|k-1} \bar{\boldsymbol{\mathcal{H}}}_{k}^{\top} (\bar{\boldsymbol{\mathcal{H}}}_{k} \boldsymbol{\mathcal{P}}_{k|k-1} \bar{\boldsymbol{\mathcal{H}}}_{k}^{\top} + \mathbf{R}_{k})^{-1} .
\label{equ:alg_gain_update}
\end{equation}
Note that $\boldsymbol{\mathcal{K}}_{k}$ represents the information gain along the transformed coordinates. By utilizing the information gain matrix $\boldsymbol{\mathcal{K}}_{k}$, we can obtain the estimation of the error-state expressed in the transformed coordinates as below
\begin{equation}
\tilde{\boldsymbol{z}}_{k|k} = \boldsymbol{\mathcal{K}}_{k} (\boldsymbol{y}_{k} - \mathbf{h}(\hat{\boldsymbol{x}}_{k|k-1})) .
\label{equ:alg_state_update}\end{equation}

\subsubsection{Transformation}

At this step, we are to convert the estimation results into the original coordinates by using the coordinate transformation \eqref{equ:T_ekf}. The state transformation is conducted by choosing for $\hat{\boldsymbol{x}}_{k|k}$ the values of $\boldsymbol{p}_{i}$ and $\psi_{i}$ such that
\begin{equation}
\hat{\boldsymbol{x}}_{k|k} = \hat{\boldsymbol{x}}_{k|k-1} + (\mathbf{T}_{k})^{-1} \tilde{\boldsymbol{z}}_{k|k} .
\label{equ:alg_state_transform}
\end{equation}
By solving the equation \eqref{equ:alg_state_transform}, we can obtain an estimate of the state expressed in the original coordinates. To derive the update equation, we first expand the error-state vector $\tilde{\boldsymbol{z}}_{k|k}$ as follows
$$
\tilde{\boldsymbol{z}}_{k|k}
=
\left[
\setlength{\arraycolsep}{2.0pt}
\begin{array}{ccccccc}
\tilde{\boldsymbol{z}}^{\boldsymbol{p}}_{1, k|k} & \tilde{\boldsymbol{z}}^{\psi}_{1, k|k} & \cdots & \tilde{\boldsymbol{z}}^{\boldsymbol{p}}_{n-1, k|k} & \tilde{\boldsymbol{z}}^{\psi}_{n-1, k|k} & \tilde{\boldsymbol{z}}^{\psi}_{n, k|k} & \tilde{\boldsymbol{z}}^{\boldsymbol{p}}_{n, k|k} \\
\end{array}
\right] .
$$
Then the orientation and position update equation for each robot can be written as
\begin{equation}
\begin{aligned}
\hat{\psi}_{i, k|k}
&= 
\begin{cases}
\hat{\psi}_{i, k|k} + \tilde{\boldsymbol{z}}_{n, k|k}^{\psi} & (i = 1) \\
\hat{\psi}_{i, k|k} + \tilde{\boldsymbol{z}}_{i-1, k|k}^{\psi} + \tilde{\boldsymbol{z}}_{n, k|k}^{\psi} & (i > 1) \\
\end{cases} \\
\hat{\boldsymbol{p}}_{i, k|k} 
&= 
\begin{cases}
\Xi \cdot (\hat{\boldsymbol{p}}_{i, k|k-1} + \tilde{\boldsymbol{z}}_{n, k|k}^{\boldsymbol{p}}) & (i = 1) \\
\Xi \cdot (\hat{\boldsymbol{p}}_{i, k|k-1} + \tilde{\boldsymbol{z}}_{n, k|k}^{\boldsymbol{p}} - \tilde{\boldsymbol{z}}_{i-1, k|k}^{\boldsymbol{p}}) & (i > 1)
\end{cases} 
\end{aligned}
\label{equ:state_update}
\end{equation}
where $\Xi = (\mathbf{I}_3 - \tilde{\boldsymbol{z}}_{n, k|k}^{\psi} \mathbf{J})^{-1}$.

Once obtaining the state estimates in the original coordinates, the coordinate transformation matrix $\mathbf{T}_{k}$ is immediately updated using $\hat{\boldsymbol{x}}_{k|k}$ according to \eqref{equ:T_ekf}.
Then the covariance transformation is conducted by using $\mathbf{T}_{k}$
\begin{equation}
\mathbf{P}_{k|k} = (\mathbf{T}_{k})^{-1} \boldsymbol{\mathcal{P}}_{k|k} ((\mathbf{T}_{k})^{-1})^\top
\label{equ:alg_cov_transform}
\end{equation}
where $\mathbf{P}_{k|k}$ is the calculated covariance estimate for the original state.

\section{Simulation Results}
\label{sec:sim}

In this section, we conduct Monte Carlo simulations to validate the proposed KD-EKF CL algorithm. In particular, we compare the performance of the KD-EKF algorithm with the other three estimators: (i) the standard EKF, (ii) the first-estimates-Jacobian (FEJ) EKF \cite{B7}, (iii) the observability-constrained (OC) EKF \cite{B6}. To achieve a comprehensive performance evaluation, two metrics are adopted: the root mean squared error (RMSE) and the normalized estimation error squared (NEES) \cite{B2}. For a consistent estimator, the NEES should have values close to the dimension of the state vector. The larger the deviation of the NEES from the dimension value, the worse the consistency.



We consider a cooperative localization scenario where a team of four robots move following helical trajectories in a $\rm 10 m \times 10 m \times 10 m$ area. The initial positions of the robots are placed randomly. The standard deviation of the linear and angular velocity measurements of each robot are $\sigma_{\boldsymbol{\nu}} = {\rm 0.3m/s}$ and $\sigma_{\varpi} = {\rm 0.08rad/s}$, respectively. The standard deviation of the relative position measurements are set to ${\rm 0.1m}$. For simplicity, we assume that all measurements occur at every time step. The simulation step is set to be $\delta t = {\rm 0.1s}$ and $100$ simulation trials are implemented. In order to make a fair comparison, we implement all the estimators using the same parameters.

\begin{figure}[!htp]
\centering
\subfloat[$3\sigma$ bounds of position]{
\includegraphics[scale=0.3, trim=10 0 40 30,clip]{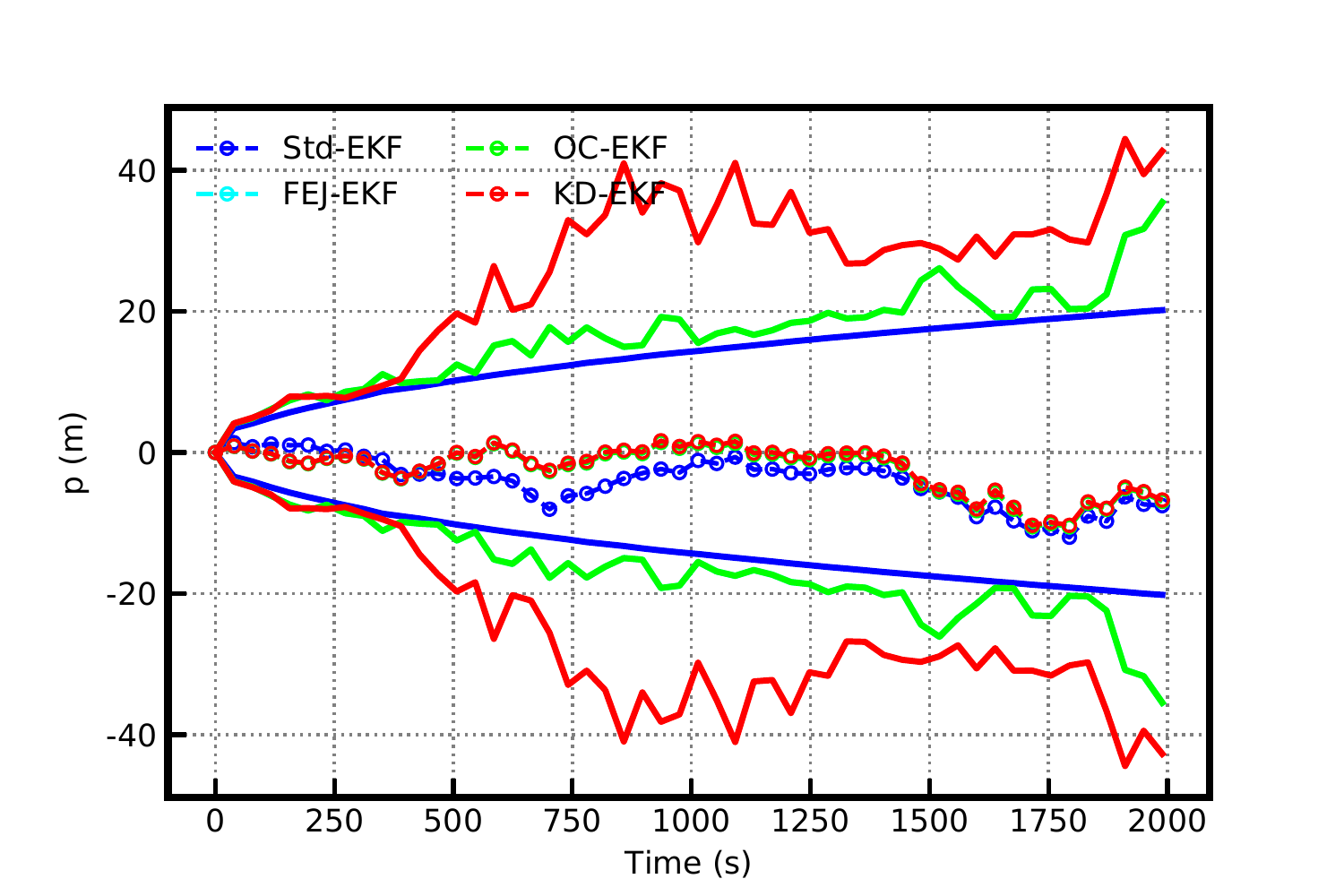}
}
\subfloat[$3\sigma$ bounds of orientation]{
\includegraphics[scale=0.3, trim=10 0 40 30,clip]{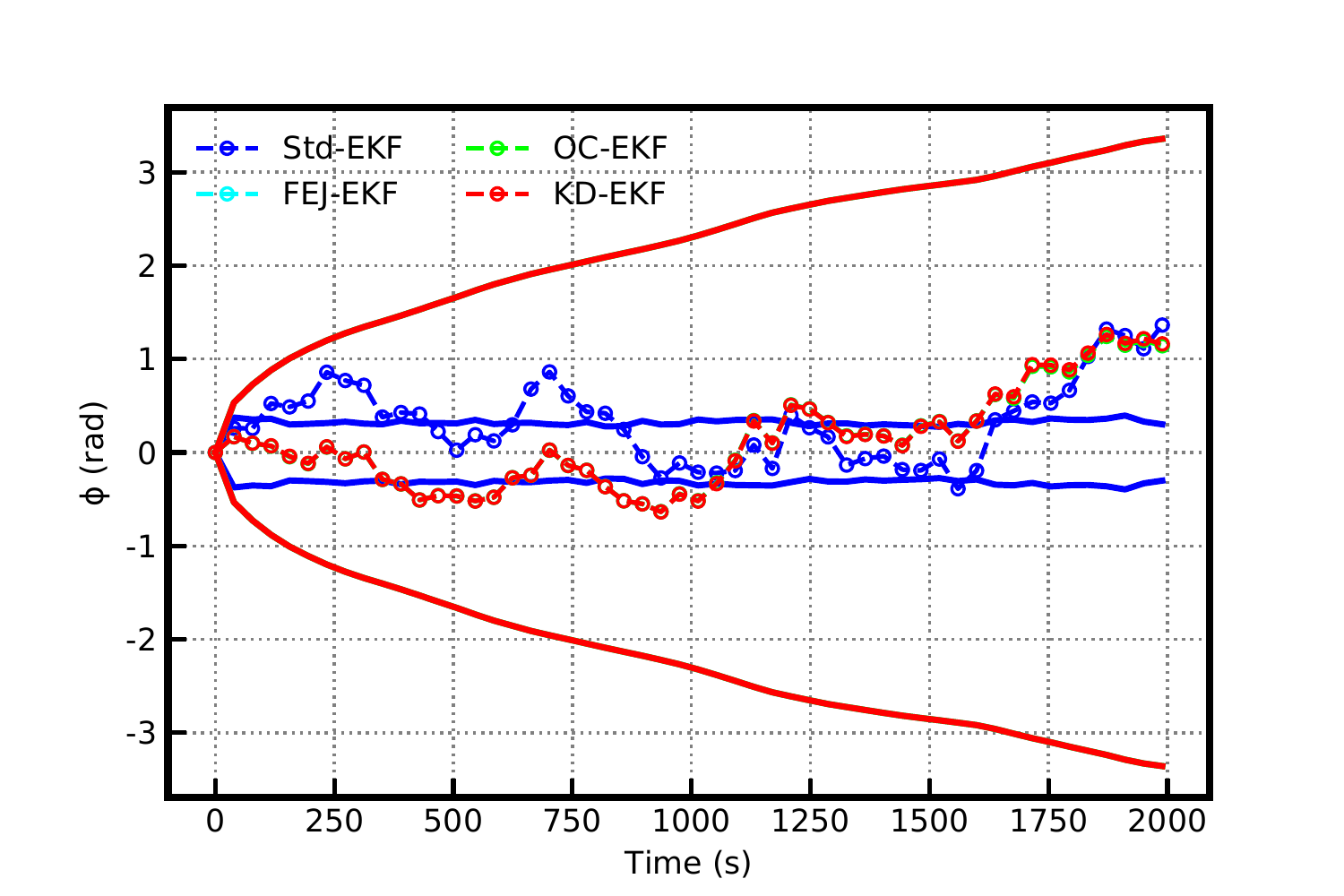}
} \\
\subfloat[RMSE]{
\includegraphics[scale=0.3, trim=10 0 40 30,clip]{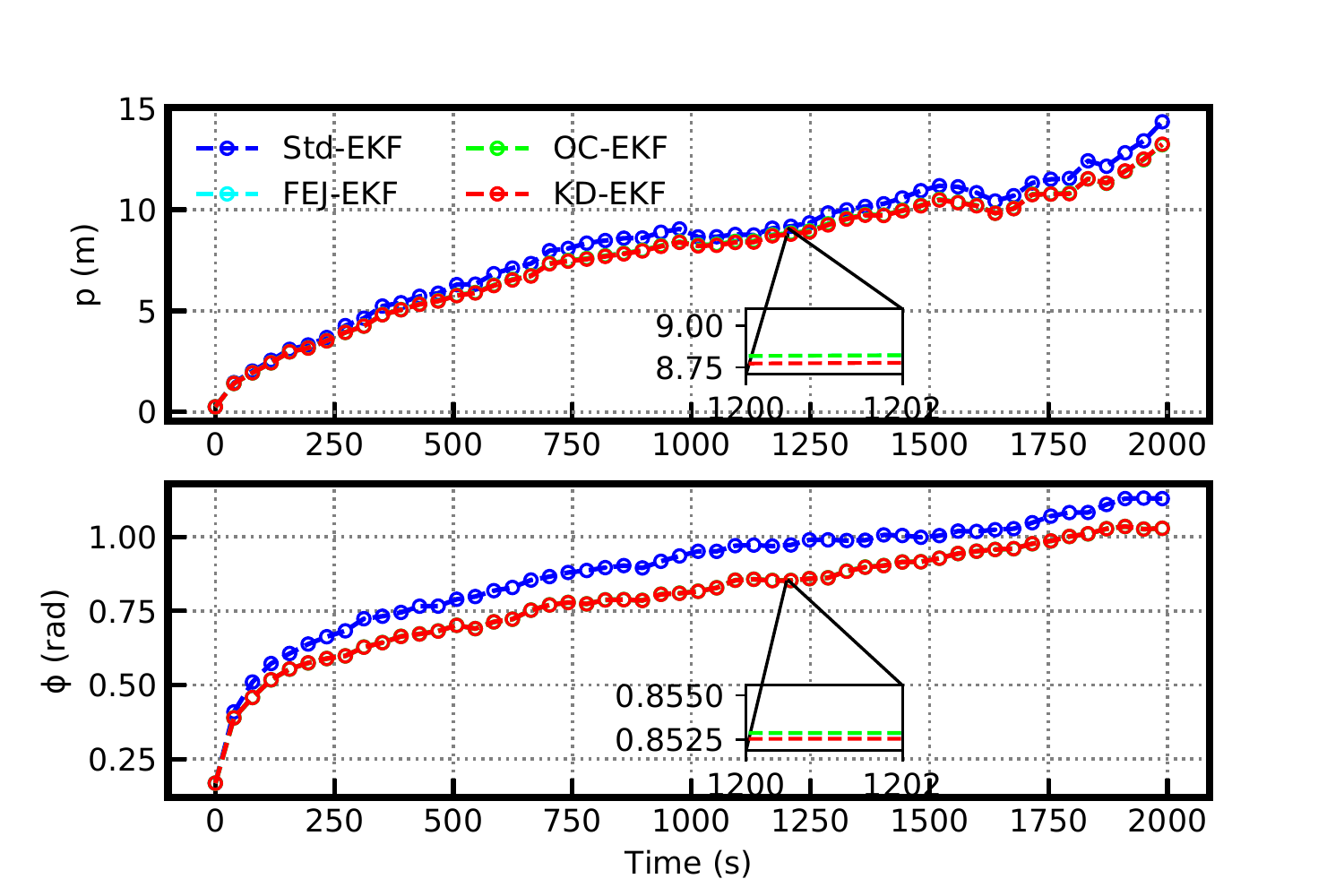}
} 
\subfloat[NEES]{
\includegraphics[scale=0.3, trim=10 0 40 30,clip]{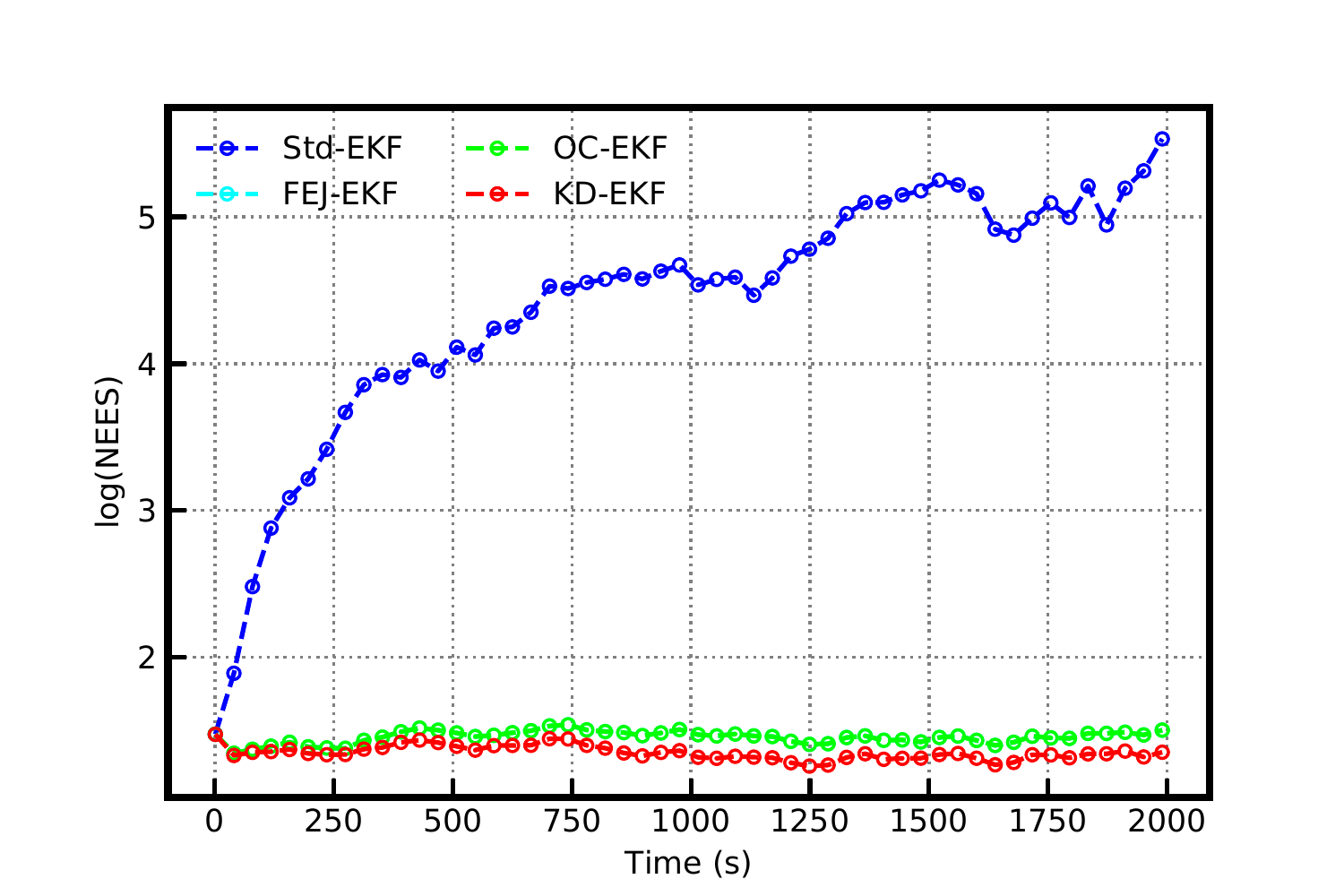}
} 
\caption{Monte Carlo simulation results for a CL scenario: (a) The position estimation error (line with markers) and $3\sigma$ bounds (line without markers); (b) The orientation estimation error and $3\sigma$ bounds; (c) The averaged position and orientation RMSE for four robots; (d) The averaged NEES for four robots. Note that the estimation results of the FEJ and OC algorithms are almost identical, which makes the corresponding lines difficult to distinguish.}
\label{fig:rmse}
\end{figure}

Fig. \ref{fig:rmse} (a) and (b) display the estimation error and corresponding $3\sigma$ bounds of uncertainty obtained from one typical run of the Monte Carlo simulation. It can be noticed that for all estimators the position estimation errors are enveloped by the corresponding $3\sigma$ bounds. However, as demonstrated in Fig. \ref{fig:rmse} (b), the standard EKF becomes inconsistent in orientation estimation since the orientation errors exceed the corresponding $3\sigma$ bounds. Evidently, the standard EKF underestimates the orientation uncertainty.

\begin{table}[!htbp]
\centering
\caption{Average RMSE and NEES over Monte Carlo simulation trials.}
\setlength{\abovecaptionskip}{0cm}
\setlength{\belowcaptionskip}{0.2cm}
\setlength{\tabcolsep}{6pt}
\renewcommand{\arraystretch}{1.2}
\begin{tabular}{cccccc} 
\toprule[2pt]  
Estimator & RMSE Pos. (m) & RMSE Ori. (rad) & NEES & \\ 
\toprule[2pt]  
Std-EKF & 8.83883 & 0.90675 & 104.90719  \\
\hline
FEJ-EKF & 8.28165 & 0.81421 & 4.29095  \\ 
\hline
OC-EKF & 8.28159 & 0.81422 & 4.29098 \\
\hline
KD-EKF & \textbf{8.26347} & \textbf{0.81405} & \textbf{3.85230} \\
\bottomrule[2pt]
\end{tabular}
\label{table:rmse}
\end{table}

Fig.~\ref{fig:rmse} (c) and (d) show the position and orientation RMSE as well as the NEES of each robot over the Monte Carlo simulations. Additionally, we summarize the average position and orientation RMSE as well as the NEES of each robot in Table~\ref{table:rmse}. It should be noted that in our case, for a consistent estimator, the NEES of each robot should have values close to $4$. Apparently, it can be observed that among these algorithms, the FEJ-EKF, OC-EKF and KD-EKF CL algorithms are superior to the standard EKF in terms of both accuracy and consistency. In addition, the KD-EKF CL algorithm performs better than the FEJ-EKF and OC-EKF with a small margin improvement. This is because the KD-EKF precisely eliminates the items causing the shrinking of the unobservable subspace.

\section{Experimental Results}
\label{sec:exp}

\begin{figure}[!htb]
\centering
\includegraphics[scale=0.1]{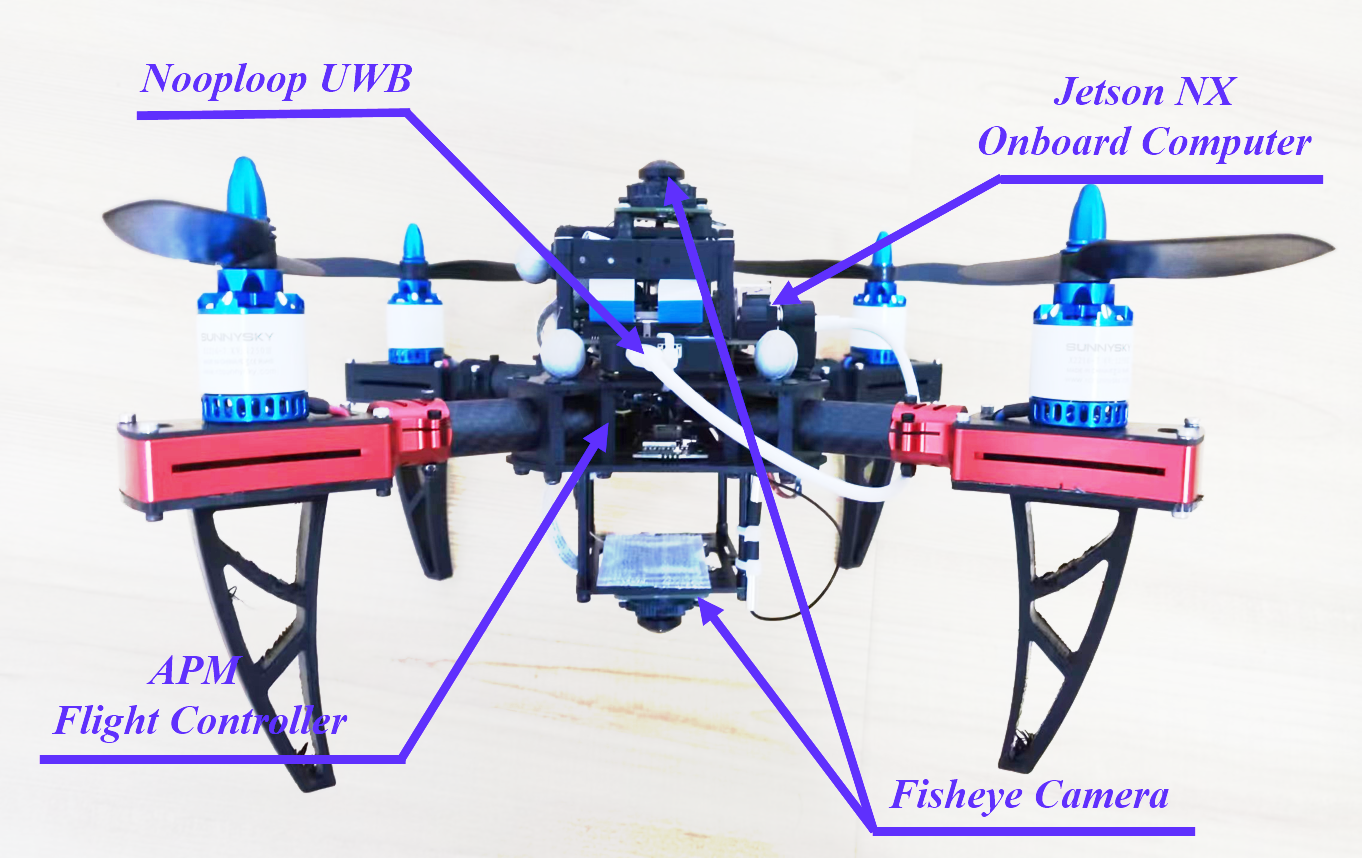}
\caption{The self-developed aerial robot in the multi-robot system, which is equipped with dual binocular fisheye cameras, Nooploop UWB modules, an APM flight controller and a Jetson NX onboard computer.}
\label{fig:exp_setup}
\end{figure}

To further test the KD-EKF algorithm, we design a multi-robot system composed of three self-developed aerial robots Hunter 2.0. The detailed configuration of the aerial robots is shown in Fig.~\ref{fig:exp_setup}. Each robot equips with an APM flight controller, a Jetson NX onboard computer, a Nooploop UWB module, and two fisheye cameras. The onboard UWB module provides relative range observations. The fisheye cameras are employed to sense the aerial robots. In particular, a YOLOv5-based detector is implemented to detect the aerial robots and attain their relative bearing observations. The detection algorithm runs in the onboard computer at 2Hz. In addition, we adopt visual-inertial odometry to sense the linear and angular velocities. To evaluate the localization results, we use the Norkov motion capture system to track the poses of these robots as ground truth.

\begin{figure}[!htb]
\centering
\subfloat[]{
\includegraphics[scale=0.5]{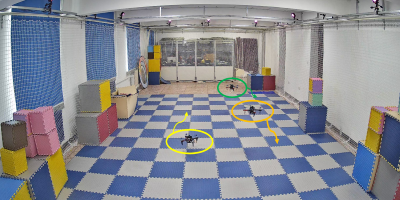}
}
\subfloat[]{
\includegraphics[scale=0.5]{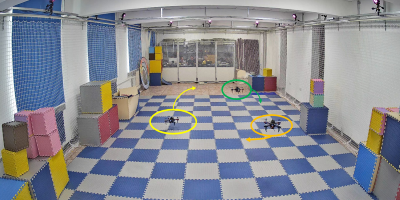}
} \\
\subfloat[]{
\includegraphics[scale=0.5]{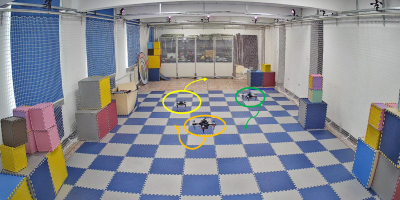}
}
\subfloat[]{
\includegraphics[scale=0.5]{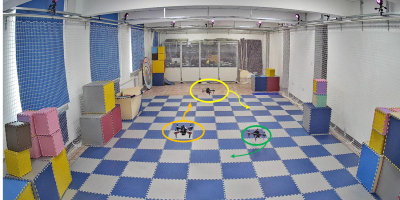}
}
\caption{Snapshots of the multi-robot system performing CL.}
\label{fig:exp_snap}
\end{figure}

To fully evaluate the performance of the KD-EKF CL algorithm, a total of four flight trials are conducted in an indoor scenario as shown in Fig. \ref{fig:exp_snap}. During the experiments, the aerial robots are maneuvered manually in a random pattern. The initial positions of the aerial robots are placed arbitrarily. Specifically, in the first two flight experiments, the aerial robots move along circular trajectories with the orientation of each robot maintaining unchanged, while in the last two flight tests, the aerial robots perform more aggressive maneuvers with the orientation changing frequently. Due to the limitation of batteries, each flight experiment can last only about $15$ minutes. Actually, the KD-EKF CL algorithm will have advantages in accuracy and consistency over long-time running.

We test the KD-EKF CL algorithm using the collected four experimental datasets and compare the performance with FEJ-EKF and OC-EKF as well as the standard EKF. Table \ref{table:exp_rmse} summarizes the average RMSE and NEES of all estimators over the four experiments. It can be observed that in the first two flight trials, the KD-EKF algorithm achieves accuracy and consistency levels similar to the FEJ-EKF and OC-EKF, while significantly outperforms the standard EKF. During the last two flight tests in which the robots perform more aggressive maneuvers, the KD-EKF algorithm still performs well and has almost the best RMSE and NEES. This shows that the KD-EKF algorithm is well-behaved for different maneuvering modes, especially for self-rotation in the last two trials. 



\begin{table}[!t]
\centering
\caption{Average RMSE and NEES for real-word experiments.}
\setlength{\abovecaptionskip}{0cm}
\setlength{\belowcaptionskip}{0cm}
\setlength{\tabcolsep}{4pt}
\renewcommand{\arraystretch}{1.2}
\begin{tabular}{cccccc} 
\toprule[2pt]  
Dataset & Metrics & Std-EKF & FEJ-EKF & OC-EKF & \textbf{KD-EKF} \\ 
\hline
& RMSE Ori. (rad) & 0.46922 & \textbf{0.43585} & 0.43586 & 0.43592 \\ 
$1$  & RMSE Pos. (m) & 7.81279 & 7.81400 & 7.81400 & \textbf{7.81067} \\
 & NEES & 13.25538 & 5.45132 & 5.45134 & \textbf{5.44064} \\
\midrule[1pt]
& RMSE Ori. (rad) & 0.41764 & 0.31303 & \textbf{0.31302} & 0.31431 \\
2 & RMSE Pos. (m) & 7.00253 & 6.94032 & 6.94038 & \textbf{6.90218} \\
& NEES & 10.33622 & 4.11743 & 4.11747 & \textbf{4.08887} \\
\midrule[1pt]
 & RMSE Ori. (rad) & 1.16232 & 0.78473 & 0.78477 & \textbf{0.78398} \\ 
3 & RMSE Pos. (m) & 10.84805 & 10.88593 & 10.88593 & \textbf{10.78341} \\ 
 & NEES &  48.95482 & 4.38647 & 4.38659 & \textbf{4.30854} \\ 
\midrule[1pt]
& RMSE Ori. (rad) & \textbf{1.00832} & 1.10165 & 1.10168 & 1.10234 \\ 
4 & RMSE Pos. (m) & \textbf{10.14009} & 10.22308 & 10.22311 & 10.17290 \\ 
 & NEES & 34.07438 & 5.45582 & 5.45595 & \textbf{5.42101} \\ 
\bottomrule[2pt]
\end{tabular}
\label{table:exp_rmse}
\end{table}

\section{Conclusions}
\label{sec:conclusion}
In this paper, we have tackled the inconsistency problem inherent in EKF-based cooperative localization (CL) by adopting a novel approach based on system decomposition. By explicitly isolating the key factors leading to the reduction in the dimension of the unobservable subspace in the Jacobians of the Kalman observable canonical form, we gained valuable insights into the fundamental cause of the issue. Based on these insights, we propose a novel cooperative localization algorithm termed KD-EKF. By performing state estimation in transformed coordinates, we effectively eliminated the influences of the identified factors, ensuring correct observability properties. Consequently, we were able to eliminate the inconsistency caused by the dimension mismatch of the unobservable subspace. Extensive simulation and experimental tests verify that the proposed algorithm performs better than the state-of-the-art CL algorithms in terms of accuracy and consistency.


Currently, we only focus on the KD-EKF applied in the cooperative localization problem. In the future, we will extend the consistency improvement algorithm to SLAM problems.

\bibliographystyle{unsrt}
\bibliography{mybibfile}

\end{document}